%% file: ArXiv_glasses.tex
\documentclass[twoside]{article}
\usepackage{aistats2016}

\usepackage{booktabs}
\usepackage{natbib}

\usepackage{hyperref}
\usepackage{url}
\usepackage[english]{babel}
\usepackage{natbib}
\usepackage{nicefrac}
\usepackage{algorithm}
\usepackage{algorithmic}
\usepackage{graphics}
\usepackage{graphicx}
\usepackage{tikz}
\usepackage{color,soul,amsmath}
\usepackage{amssymb}
\usetikzlibrary{bayesnet}
\tikzstyle{connect}=[-latex]
\tikzstyle{allconnected}=[line width=0.1cm]

\newcommand{\I}{\mathcal{I}}
\newcommand{\ud}{\mathrm{d}}
\newcommand{\E}{\mathbb{E}}

\newcommand{\GP}{\mathcal{GP}}
\newcommand{\N}{\mathcal{N}}

\newcommand{\future}{\mathcal{F}}
\newcommand{\IR}{\mathbb{R}}

\newcommand{\reals}{\mathbb{R}}

\newcommand{\xst}{x_{\ast}}
\newcommand{\yst}{y_{\ast}}

\usepackage{xspace}
\newcommand{\acr}[1]{\textsc{#1}\xspace}
\newcommand{\gp}{\acr{gp}}
\newcommand{\dpp}{\acr{dpp}}
\newcommand{\us}{\acr{glasses}}
\newcommand{\direct}{\acr{direct}}
\newcommand{\lbfgs}{\acr{l-bfgs}}
\newcommand{\map}{\acr{map}}
\newcommand{\ep}{\acr{ep}}
\newcommand{\bo}{\acr{bo}}
\newcommand{\mpi}{\acr{mpi}}
\newcommand{\el}{\acr{el}}
\newcommand{\lcb}{\acr{gp-lcb}}

\newcommand*{\addheight}[2][.5ex]{%
  \raisebox{0pt}[\dimexpr\height+(#1)\relax]{#2}%
}

\input{notationDef.tex}

\begin{document}

%

%

\twocolumn[

\aistatstitle{GLASSES: Relieving The Myopia Of Bayesian Optimisation}

\aistatsauthor{Javier Gonz\'alez \And Michael Osborne \And Neil D. Lawrence}

\aistatsaddress{
University of Sheffield\\ 
Dept. of Computer Science \& \\
Chem. and Biological Engineering\\
j.h.gonzalez@sheffield.ac.uk\\
 \And 
 University of Oxford \\
Dept. of Engineering Science\\
 mosb@robots.ox.ac.uk
 \And University of Sheffield\\
 Dept. of Computer Science\\
 n.lawrence@sheffield.ac.uk } ]

\begin{abstract}
    We present \us: Global optimisation with Look-Ahead through Stochastic Simulation and Expected-loss Search. 
    The majority of global optimisation approaches in use are myopic, in only considering the impact of the next function value; the non-myopic approaches that do exist are able to consider only a handful of future evaluations. 
    Our novel algorithm, \us, permits the consideration of dozens of evaluations into the future.  This is done by approximating the ideal \emph{look-ahead} loss function, which is expensive to evaluate, by a cheaper alternative in which the future steps of the algorithm are simulated beforehand. An Expectation Propagation algorithm is used to compute the expected value of the loss.
    We show that the far-horizon planning thus enabled leads to substantive performance gains in empirical tests. 
\end{abstract}
\section{Introduction} 
\label{sec:introduction}

\begin{figure*}[t!]
\centering
\begin{tikzpicture}

    \node[obs] (D0) {$\dataSet_0$};
    \node[latent, right=of D0, xshift=1.2cm] (D1) {$\dataSet_1$};
    \node[draw=none, right=of D1, xshift=1.2cm] (Ddots) {$\ldots$};
    \node[latent, right=of Ddots, xshift=1.2cm] (Dn) {$\dataSet_n$};

    \node[det, below=of D0, xshift=1.2cm] (xst) {$\xst$};
    \node[latent, right=of xst, xshift=1.2cm] (x2) {$x_2$};
    \node[latent, right=of x2, xshift=4.8cm] (xn) {$x_n$};

    \node[latent, below=of xst] (yst) {$\yst$};
    \node[latent, below=of x2] (y2) {$y_2$};
    \node[draw=none, right=of y2, xshift=1.2cm] (ydots) {$\ldots$};
    \node[latent, below=of xn] (yn) {$y_n$};

    \path 
        (D0) edge [connect] (D1)
        (D0) edge [connect] (xst)
        (xst) edge [connect] (yst)
        (xst) edge [connect] (D1)
        (yst) edge [connect, bend right=20] (D1)
        (yst) edge [allconnected] (y2)

        (D1) edge [connect] (Ddots)
        (D1) edge [connect] (x2)
        (x2) edge [connect] (y2)
        (x2) edge [connect] (Ddots)
        (y2) edge [connect, bend right=20] (Ddots)
        (y2) edge [allconnected] (ydots)

        (D1) edge [connect] (Ddots)
        (D1) edge [connect] (x2)
        (x2) edge [connect] (y2)
        (x2) edge [connect] (Ddots)
        (y2) edge [connect, bend right=20] (Ddots)
        (ydots) edge [allconnected] (yn)

        (Ddots) edge [connect] (Dn)
        (Dn) edge [connect] (xn)
        (xn) edge [connect] (yn)        
        ;
\end{tikzpicture}
\caption{
    A Bayesian network describing the $n$-step lookahead problem. The shaded node ($\dataSet_0$) is known, and the diamond node ($\xst$) is the current decision variable. All $y$ nodes are correlated with one another under the \gp model. Note that the nested maximisation problems required for $x_i$ and integration problems required for $y_*$ and $y_i$ (in either case for $i=2,\ldots, n$) render inference in this model prohibitively computationally expensive.
}
\label{fig:bayes_net}
\end{figure*}
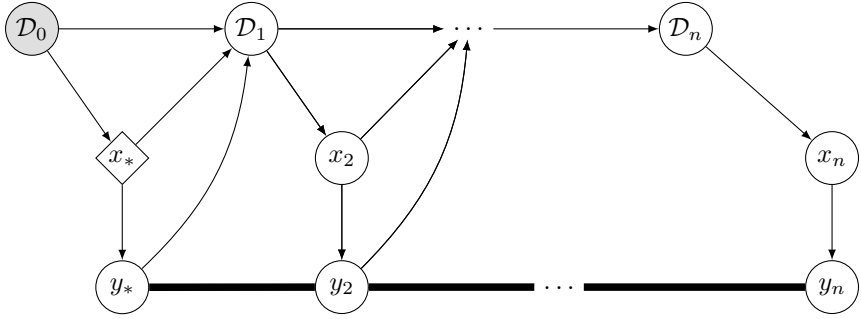

Global optimisation is core to any complex problem where design and choice play a role. 
Within Machine Learning, such problems are found in the tuning of hyperparameters \citep{Snoek*Larochelle*Adams_2012}, sensor selection \citep{Garnett*Osborne*Roberts_2010} or experimental design \citep{gonzalez2014, martinez-cantin_bayesian_2009}. 
Most global optimisation techniques are myopic, in considering no more than a single step into the future. 
Relieving this myopia requires solving the \emph{multi-step lookahead} problem: the global optimisation of an function by considering the significance of the next function evaluation on function evaluations (steps) further into the future. 
It is clear that a solution to the problem would offer performance gains.
For example, consider the case in which we have a budget of two evaluations with which to optimise a function $f(x)$ over the domain $\inputSpace = [0, 1] \subset \reals$. 
If we are strictly myopic, our first evaluation will likely be at 
$x=\nicefrac{1}{2}$, and our second then at only one of $x=\nicefrac{1}{4}$ and $x=\nicefrac{3}{4}$. 
This myopic strategy will thereby result in ignoring half of the domain $\inputSpace$, regardless of the second choice. 
If we adopt a two-step lookahead approach, we will select function evaluations that will be more evenly distributed across the domain by the time the budget is exhausted. 
We will consequently be better informed about $f$ and its optimum.

There is a limited literature on the multi-step lookahead problem.
\cite{osborne_gaussian_2009} perform multi-step lookahead by optimising future evaluation locations, and sampling over future function values. 
This approach scales poorly with the number of future evaluations considered, and the authors present results for no more than two-step lookahead.
\citep{Marchant*Ramos*Sanner*2014} reframe the multi-step lookahead problem as a partially observed Markov decision process, and adopt a Monte Carlo tree search approach in solving it. 
Again, the scaling of the approach permits the authors to consider no more than six steps into the future. In the past, the multi-step look ahead problem was studied by \cite{StreltsovVakili1999} proposing a utility function that maximizes the total `reward' of the algorithm by taking into account the cost of future computations, rather than trying to find the optimum after a fixed number of evaluations.

Interestingly, there exists a link between the multi-step lookahead problem and \emph{batch} Bayesian optimisation \citep{Ginsbourger2009,Azimi2012}. In this later case, batches of locations rather than individual observations are selected in each iteration of the algorithm and evaluated in parallel. When such locations are selected \emph{greedily}, that is, one after the other, the key to selecting good batches relies on the ability of the batch criterion of predicting future steps of the algorithm. In this work we will exploit this parallelism to compute a non-myopic loss for Bayesian optimisation. 


This paper is organised as follows. In Section \ref{sec:background} we formalise the problem and describe the contributions of this work. Section \ref{sec:glasses} describe the details of the proposed algorithm. Section \ref{sec:experiments} illustrates the superior performance of \us in a variety of test functions and we conclude in Section \ref{sec:conclusions} with a discussion about the most interesting results observed in this work.

\section{Background and challenge}\label{sec:background}
\subsection{Bayesian optimisation with one step look-ahead} 
\label{sec:bayesian_optimisation}

Let $f: {\mathcal X} \to \Re$ be well behaved function defined on a compact subset ${\mathcal X} \subseteq \Re^{\inputDim}$. We are interested in solving the global optimization problem of finding 
$$\latentVector_{M} = \arg \min_{\latentVector \in {\mathcal X}} f(\latentVector).$$ 
We assume that $f$ is a \emph{black-box} from which only perturbed evaluations of the type $\dataScalar_i = f(\latentVector_i) + \noiseScalar_i$, with $\noiseScalar_i \sim\mathcal{N}(0,\dataStd^2)$, are  available. Bayesian Optimization (\bo) is an heuristic strategy to make a series of evaluations $\latentVector_1,\dots,\latentVector_n$ of $f$, typically very limited in number,  such that the the minimum of $f$ is evaluated as soon as possible \citep{Lizotte_2008,Jones_2001,Snoek*Larochelle*Adams_2012,Brochu*Cora*DeFreitas_2010}.

Assume that $N$ points have been gathered so far, having a dataset $\dataSet_0 = \{(\latentVector_i,\dataScalar_i)\}_{i=1}^N = (\latentMatrix_0,\dataVector_0)$. Before collecting any new point, a surrogate probabilistic model for $f$ is calculated. This is typically a Gaussian Process (\gp) $p(f) = \mathcal{GP}(\mu; k)$ with mean function $\mu$ and a covariance function $k$, and whose parameters will be denoted by $\parameterVector$.  Let $\I_0$ be the current available information: the conjunction of $\dataSet_0$, the model parameters and the model likelihood type.  Under Gaussian likelihoods, the predictive distribution for $y_*$ at $\latentVector_*$ is also Gaussian with mean posterior mean and variance
$$\mu(\latentVector_{*}|\I_0) = \kernelVector_{\parameterVector}(\latentMatrix_*)^\top[\kernelVector_{\parameterVector} + \sigma^2 \textbf{I}]^{-1}\dataVector \mbox{ and}$$
$$\sigma^2(\latentVector_*|\I_0)=k_{\parameterVector}(\latentVector_*,\latentVector_*)-\kernelVector_{\parameterVector}(\latentVector_*)^\top[\kernelMatrix_{\parameterVector}+\sigma^2 \textbf{I}]^{-1}\kernelVector_{\parameterVector}(\latentVector_*),$$
where $\kernelMatrix_{\parameterVector}$ is the matrix such that $(\kernelMatrix_{\parameterVector})_{ij}=\kernelScalar_{\parameterVector}(\latentVector_i,\latentVector_j)$,  $\kernelVector_{\parameterVector}(\latentVector_{*}) = [k_{\parameterVector}(\latentVector_1,\latentVector_{*}),\dots,k_{\parameterVector}(\latentVector_n,\latentVector_{*})]^\top$ \citep{Rasmussen:2005:GPM:1162254}.

Given the \gp model, we now need to determine the best location to sample. Imagine that we only have one remaining evaluation ($n=1$) before we need to report our inferred location about the minimum of $f$. Denote by $\eta = \min \{\dataVector_0\}$, the current best found value. We can define the loss of evaluating $f$ this last time at $\latentVector_*$ assuming it is returning $y_*$ as
$$\lambda(y_*)\triangleq \left\{ \begin{array}{lcl}
y_*;             & \mbox{if}  &  y_* \leq \eta \\
 \eta; & \mbox{if}  & y_* > \eta. \\
\end{array}
\right.$$
Its expectation is 
$$\ \Lambda_1(\latentVector_*| \I_0) \triangleq \E[ \min (y_*,\eta)]= \int \lambda(y_*)p(y_* |\latentVector_*,\I_0)\ud y_*$$
where the subscript in $\Lambda$ refers to the fact that we are considering one future evaluation.  Giving the properties of the \gp, $\Lambda_1(\latentVector_*| \I_0)$ can be computed in closed form for any $\latentVector_* \in \mathcal{X}$. In particular, for $\Phi$ the usual Gaussian cumulative distribution function, we have that
\begin{eqnarray}\label{eq:expected_myopic_loss}
\Lambda_1(\latentVector_*| \I_0) &  \triangleq & \eta \int_{\eta}^{\infty} \mathcal{N}(y_*;\mu,\sigma^2) \ud y_* \\ \nonumber
& +  & \int_{-\infty}^{\eta} y_* \mathcal{N}(y_*;\mu,\sigma^2) \ud y_*  \\ \nonumber
& = &  \eta +(\mu  - \eta) \Phi (\eta ; \mu, \sigma^2) - \sigma^2 \mathcal{N} (\eta, \mu, \sigma^2), \nonumber
\end{eqnarray}

where we have abbreviated $\sigma^2(y_*|\I_0)$ as $\sigma^2$ and $\mu(y_{*}|\I_0)$ as $\mu$. Finally, the next evaluation is located   where $\Lambda_1(\latentVector_*| \I_0) $ gives the minimum value. This point can  be obtained by any gradient descent algorithm since analytical expressions for the gradient and Hessian of $\Lambda_1(\latentVector_*| \I_0)$  exist \citep{osborne_bayesian_2010}.



\begin{figure*}[t!]
\centering
\begin{tikzpicture}

    \node[obs] (D0) {$\dataSet_0$};

    \node[det, below=of D0, xshift=1.2cm] (xst) {$\xst$};
    \node[latent, right=of xst, xshift=1.2cm] (x2) {$x_2$};
    \node[draw=none, right=of x2, xshift=1.2cm] (xdots) {$\ldots$};
    \node[latent, right=of x2, xshift=4.8cm] (xn) {$x_n$};

    \node[latent, below=of xst] (yst) {$\yst$};
    \node[latent, below=of x2] (y2) {$y_2$};
    \node[draw=none, right=of y2, xshift=1.2cm] (ydots) {$\ldots$};
    \node[latent, below=of xn] (yn) {$y_n$};

    \path 
        (D0) edge [connect] (xst)
        (xst) edge [connect] (yst)
        (yst) edge [allconnected] (y2)

        (D0) edge [connect, bend left=20] (x2)
        (D0) edge [connect, bend left=10] (xn)

        (xst) edge [allconnected] (x2)
        (x2) edge [allconnected] (x2)
        (x2) edge [allconnected] (xdots)
        (xdots) edge [allconnected] (xn)

        (x2) edge [connect] (y2)
        (y2) edge [allconnected] (ydots)

        (x2) edge [connect] (y2)
        (ydots) edge [allconnected] (yn)

        (xn) edge [connect] (yn)        
        ;
\end{tikzpicture}
\caption{
    A Bayesian network describing our approximation to the $n$-step lookahead problem. The shaded node ($\dataSet_0$) is known, and the diamond node ($\xst$) is the current decision variable, which is now directly connected with all future steps of the algorithm. Compare with Figure \ref{fig:bayes_net}: the sparser structure renders our approximation computationally tractable. 
}
\label{fig:bayes_net_glasses}
\end{figure*}
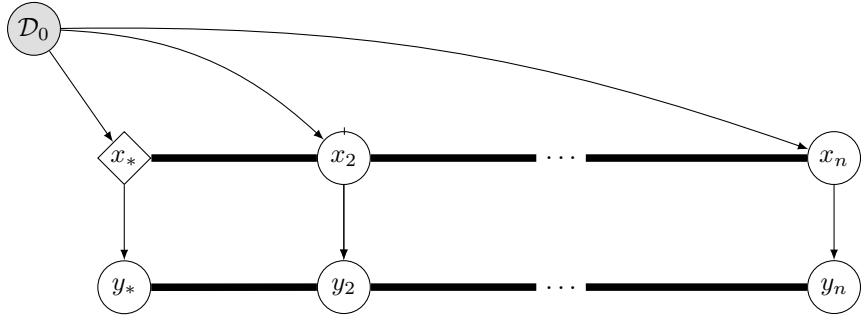

\subsection{Looking many steps ahead}
Expression~(\ref{eq:expected_myopic_loss}) can also be used as a myopic approximation to the optimal decision when $n$ evaluations of $f$ remain available. Indeed, most \bo methods are myopic and ignore the future decisions that will be made by the algorithm in the future steps. 

Let $\{(\latentVector_j,\dataScalar_j)\}$ for $j=1,\dots,n$ be the remaining $n$ available evaluations and by $\I_j$ the available information after the data set $\dataSet_0$ has been augmented with $(\latentVector_1,\dataScalar_1),\dots,(\latentVector_j,\dataScalar_j)$ and the parameters $\parameterVector$ of the model updated. We use $\Lambda_n(\latentVector_*|\I_0 )$ to denote the expected loss of selecting $\latentVector_*$ given $\I_0$ and  considering $n$ future evaluations.  A proper Bayesian formulation allows us to define this \emph{long-sight} loss  \citep{osborne_bayesian_2010}  as\footnote{We assume that $p(\latentVector_*|\I_0)=1$.}
\begin{eqnarray}\nonumber
\Lambda_n(\latentVector_*|\I_0 ) &= & \int \lambda(y_n) \prod_{j=1}^{n}p(y_{j}|\latentVector_{j},\I_{j-1}) p(\latentVector_{j}|\I_{j-1})\\\label{eq:expected_nonmyopic_loss}
& & \ud y_*\dots \ud y_n \ud\latentVector_2\dots \ud\latentVector_n
\end{eqnarray}
where 
$$p(y_{j}|\latentVector_{j},\I_{j-1})= \mathcal{N} \left(y_{j};\mu(\latentVector_{j};\I_{j-1}),\sigma^2(\latentVector_{j}|\I_{j-1} ) \right)$$ 
is the predictive distribution of the \gp at $\latentVector_{j}$  and 
$$p(\latentVector_{j}|\I_{j-1}) = \delta \bigr(\latentVector_{j} - \arg \min_{\latentVector_* \in {\mathcal X}} \Lambda_{n-j+1}(\latentVector_*|\I_{j-1})\bigl)$$ 
reflects the optimization step required to obtain $\latentVector_{j}$ after all previous the evaluations $f$ have been iteratively optimized and marginalized.  The graphical probabilistic model underlying (\ref{eq:expected_nonmyopic_loss}) is illustrated in Figure \ref{fig:bayes_net}.

To evaluate Eq.~(\ref{eq:expected_nonmyopic_loss}) we can successively sample from $y_1$ to $y_{j-1}$ and optimize for the appropriate $\Lambda_{n-j+1}(\latentVector_*|\I_{j-1})$. This is in done in \citep{osborne_bayesian_2010} for only two steps look ahead. The reason why further horizons remain unexplored is the  computational burden required to compute this loss for many steps ahead. Note that analytical expression are only available in the myopic case $\Lambda_1(\latentVector_*| \I_0)$.

\subsection{Contributions of this work}

The goal of this work is \emph{to propose a computationally efficient approximation to Eq.~(\ref{eq:expected_nonmyopic_loss}) for many steps ahead able to relieve the myopia of classical Bayesian optimization}. The contributions of this paper are:
\begin{itemize}
\item A new algorithm, \us, to relieve the myopia of Bayesian optimisation that is able to efficiently take into account dozens of steps ahead. The method is based on the prediction of the future steps of the myopic loss to efficiently integrate out a long-side loss. 
\item The key aspect of our approach is to split the recursive optimization marginalization loop in Eq.~(\ref{eq:expected_nonmyopic_loss}) into two independent optimisation-marginalization steps that jointly act on all the future steps. We propose an Expectation-Propagation formulation for the joint marginalisation and we discuss different strategies to carry out the optimisation step.
\item Together with this work, we deliver a \emph{open source Python code framework}\footnote{http://sheffieldml.github.io/GPyOpt/} containing a fully functional  implementation of the method useful to reproduce the results of this work and applicable in general global optimisation problems. As we mentioned in the introduction of this work, there exist a limited literature in \bo non-myopic methods and, to our knowledge, none available generic \bo package can be used with myopic loss functions. 
\item Simulations: New practical experiments and insights that show that non-myopic methods outperform myopic approaches in a benchmark of optimisation problems. 
\end{itemize}

\begin{table*}[t!]
\begin{tabular}{ccc}
      \addheight{\includegraphics[width=54mm]{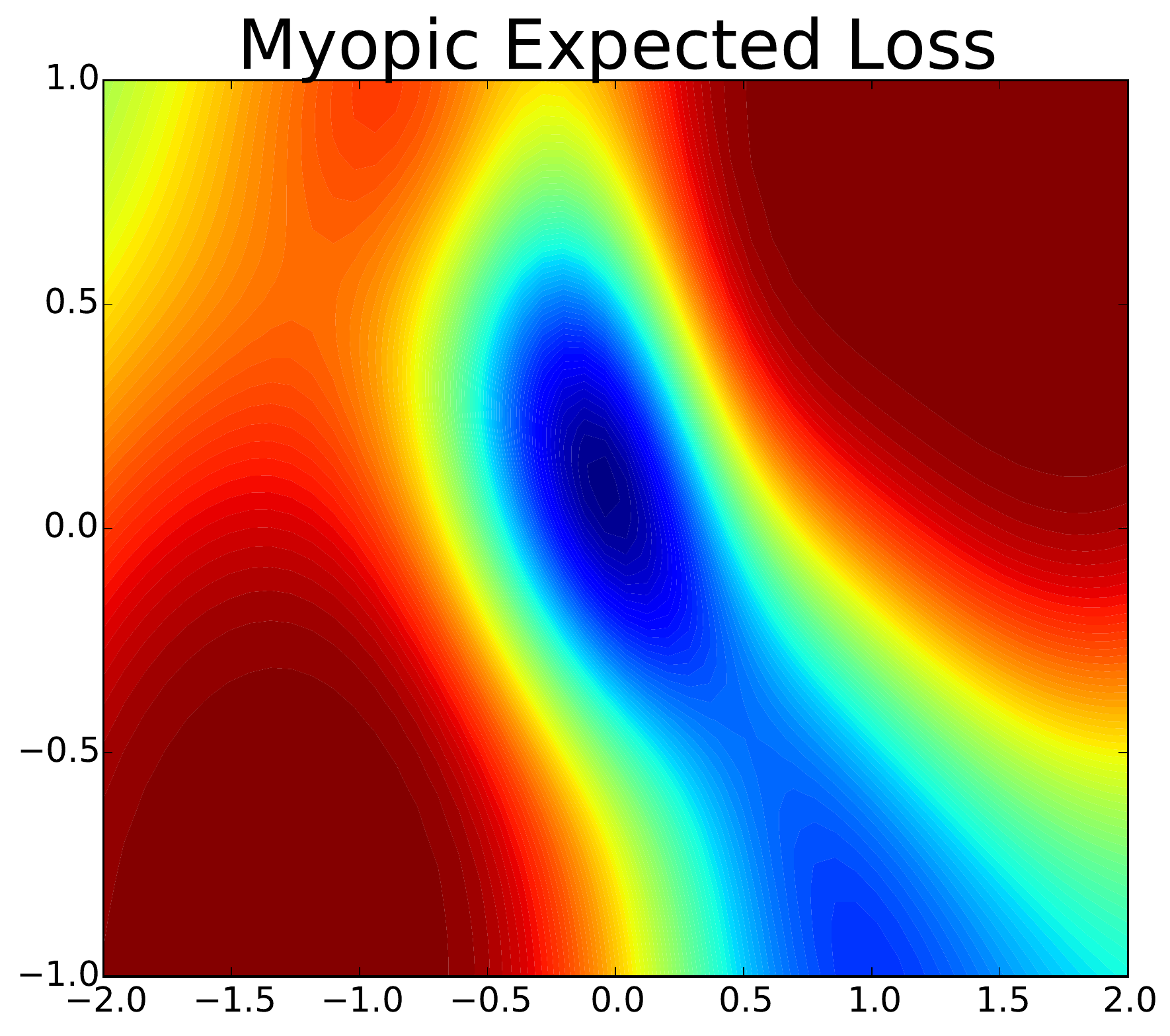}} &
      \addheight{\includegraphics[width=54mm]{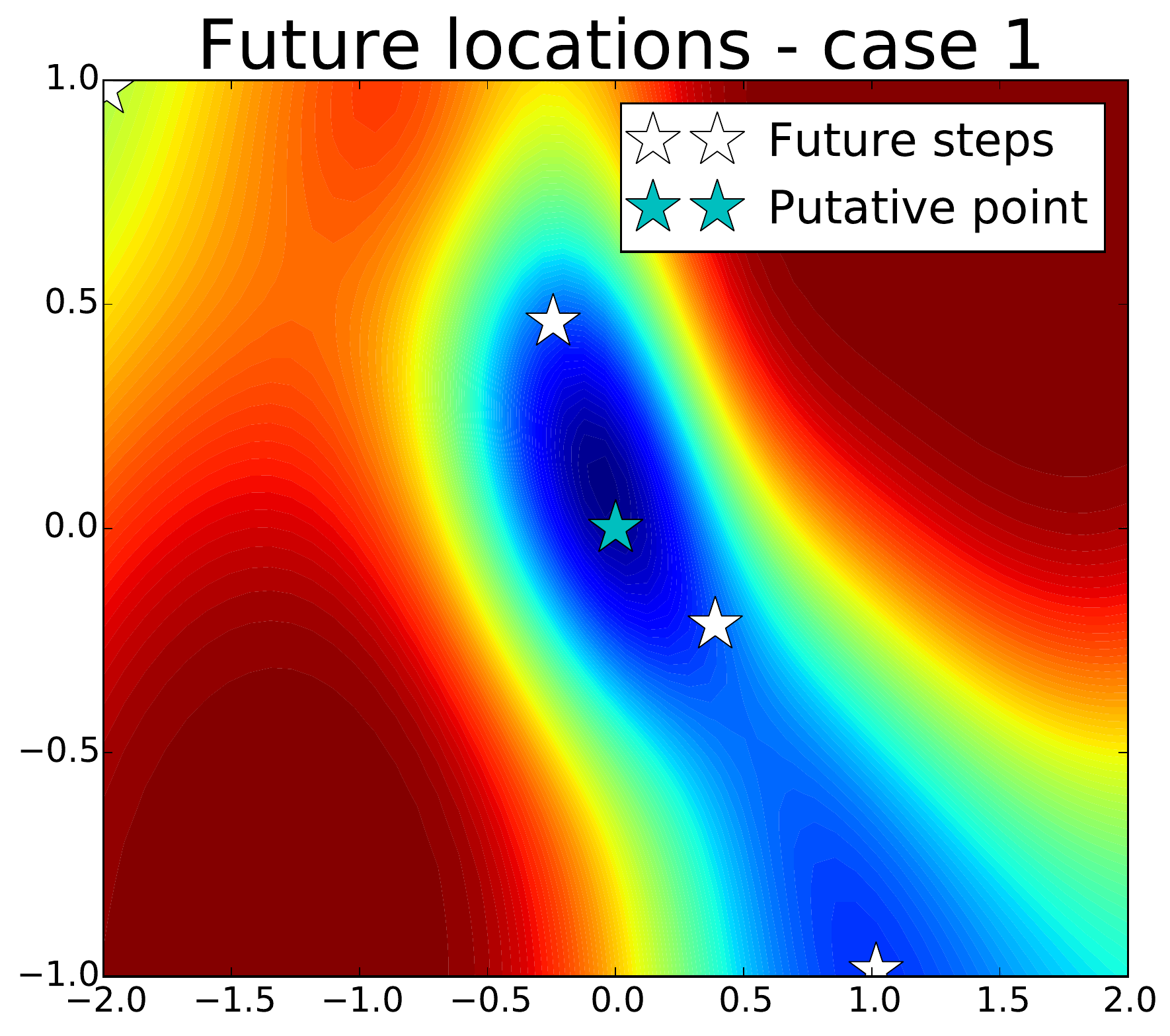}}  &
      \addheight{\includegraphics[width=54mm]{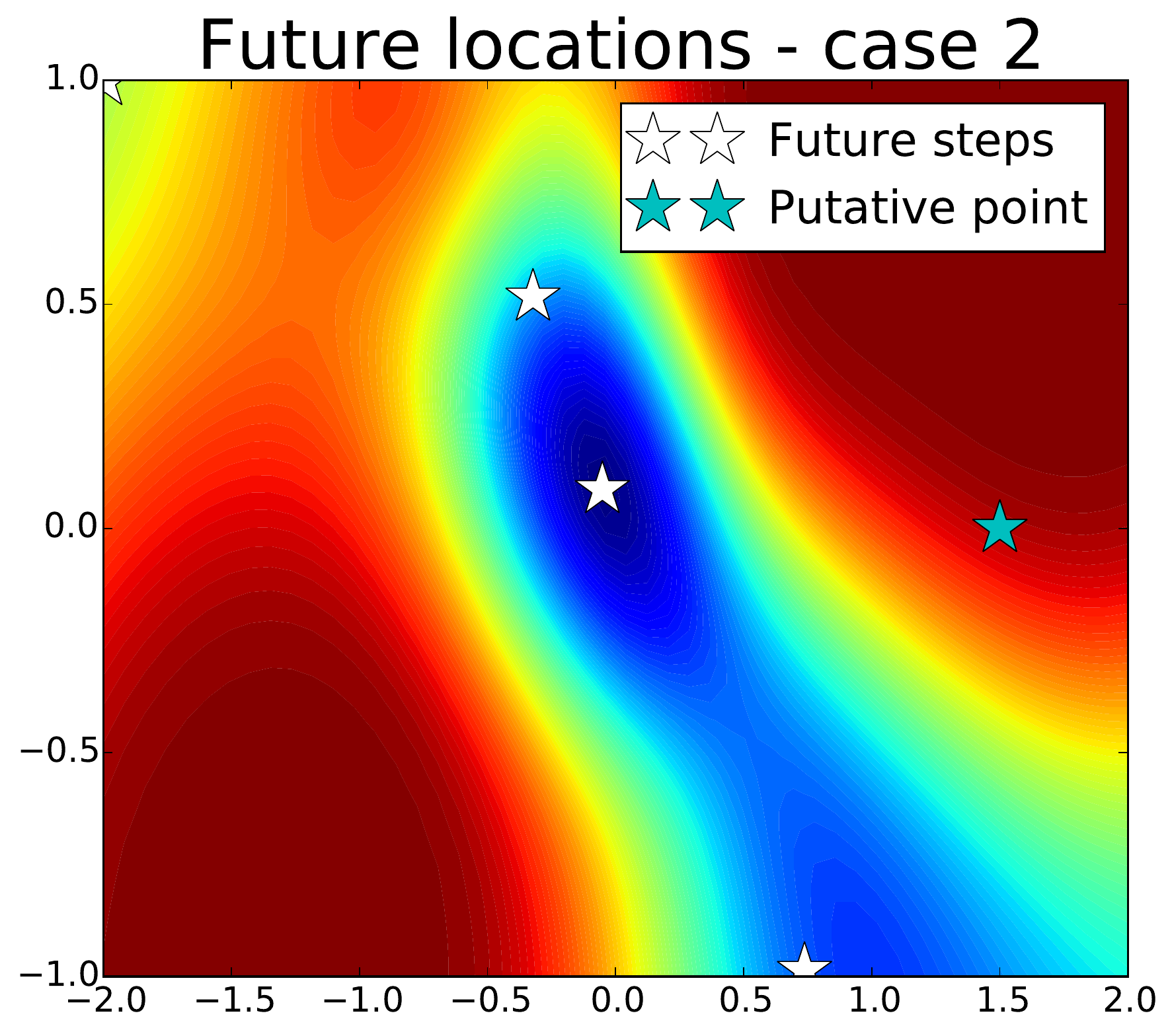}}
\end{tabular}
\end{table*}

 \section{The GLASSES algorithm}\label{sec:glasses}

\begin{figure*}[t!]
\begin{center}\label{fig:steps_ahead}
\includegraphics[width=16cm]{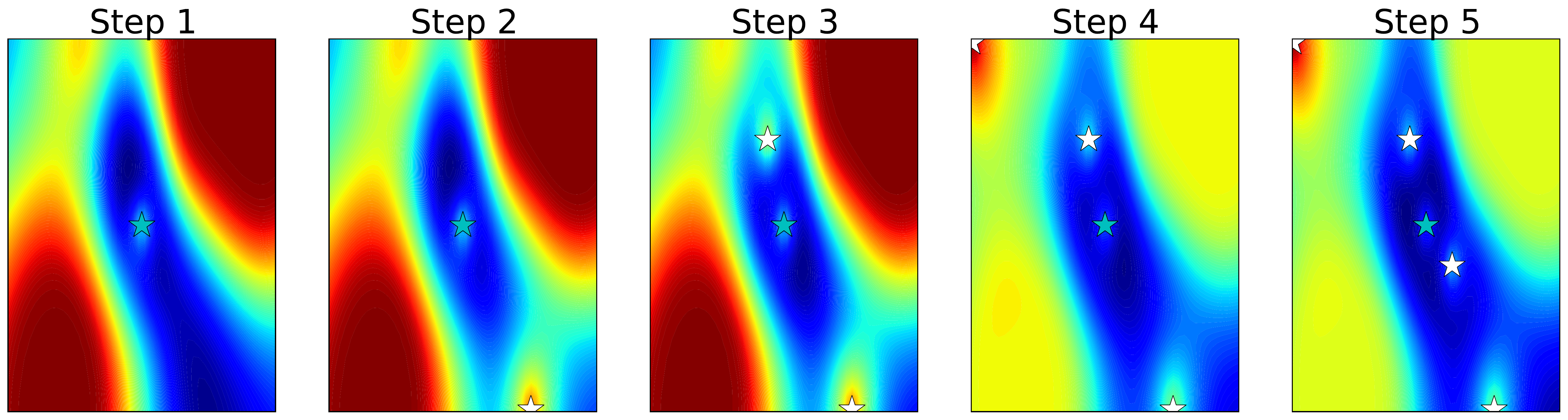}
\end{center}\caption{\emph{Top row}: (left) Myopic expected loss computed after 10 observations of the Six-Hump Camel function; (center, case 1) predicted steps ahead when the putative point, grey star, is close to the global optimum of the acquisition; (right, case 2) predicted steps ahead when the putative input is located far from the optimum of the acquisition. \emph{Bottom row}: iterative decision process that predicts that 5 steps look ahead of case 2. Every time a point is selected, the loss is penalised in a neighbourhood, encouraging the next location to be selected far from any previous location point but still in a region where the value of the loss is low.}
\end{figure*}

As detailed in the previous section, a proper multi-step look ahead loss function requires the iterative optimization-marginalization of the future steps, which is computationally intractable. A possible way of dealing with this issue is to jointly model our epistemic uncertainty over the future locations $\latentVector_2,\dots,\latentVector_n$ with a joint probability distribution  $p(\latentVector_2,\dots,\latentVector_n|\I_{0}, \latentVector_*) $ and to consider the expected loss 
\begin{equation}\label{eq:eq:expected_nonmyopic_loss2}
\Gamma_n(\latentVector_*|\I_0 ) = \int \lambda(y_n) p(\dataVector|\textbf{X},\I_{0},\latentVector_*) p(\textbf{X}|\I_{0},\latentVector_*) \ud \dataVector \ud\textbf{X} \
\end{equation}
for $\dataVector=\{y_*,\dots,\dots,y_n\}$ the vector of future evaluations of $f$ and $\textbf{X}$ the $(n-1)\times \inputDim$ dimensional matrix whose rows are the future evaluations $\latentVector_2,\dots,\latentVector_n$. $p(\dataVector|\textbf{X},\I_{0},\latentVector_*)$ is multivariate Gaussian, since it corresponds to the predictive distribution of the \gp at $\textbf{X}$. The graphical probabilistic model underlying (\ref{eq:expected_nonmyopic_loss}) is illustrated in Figure \ref{fig:bayes_net_glasses}. $\Gamma_n(\latentVector_*|\I_0 )$ differs from $\Lambda_n(\latentVector_*|\I_0)$ in the fact that all future evaluations are modelled jointly rather then sequentially. A proper choice of $p(\textbf{X}|\I_{0}, \latentVector_*)$ is crucial here. An interesting option would be to choose $p(\textbf{X}|\I_{0}, \latentVector_*)$ to be some determinantal point process (\dpp)%
\footnote{
    A determinantal point process is a probability measure over sets that is entirely characterised by the determinant of some (kernel) function.
}
defined on ${\mathcal X}$ \citep{Affandi:NIPS2013} and integrate Eq.~(\ref{eq:eq:expected_nonmyopic_loss2}) with respect to $\latentVector_2,\dots,\latentVector_n$ by averaging over multiple samples \citep{MAL-044,KuleszaT11}. 
DPPs provide a density over sample locations that induces them to be dissimilar to each other (as well-spaced samples should), but that can be concentrated in chosen regions (such as regions of low myopic expected loss).
However, although DPPs have nice computational properties in discrete sets, here we would need to take samples from the \dpp by conditioning on $\latentVector_*$ and the number of steps ahead. Although this is possible in theory, the computational burden of generating these samples will make this strategy impractical. 

An alternative and more efficient approach, that we explore here, is to work with a fixed $\textbf{X}$, which we assume it is computed beforehand. As we show in this section, although this approach does not make use of our epistemic uncertainty on the future steps, it drastically reduces the computational burden of approximating $\Lambda_n(\latentVector_*|\I_0 )$.

\subsection{Oracle multiple steps look-ahead expected loss}

Suppose that  we had access to an oracle function $\future_{n}: {\mathcal X}\rightarrow  {\mathcal X}\times \overset{n}{\cdots}\times {\mathcal X}$ able to predict the $n$ future  locations that the loss $\Lambda_n(\cdot)$ would suggest if we started evaluating $f$ at $\latentVector_*$. In other words, $\future_{n}$ takes the putative location $\latentVector_*$ as input and it returns $\latentVector_*$ and the predicted future locations  $\latentVector_2,\dots, \latentVector_n$. We work here under the assumption that the oracle has perfect information about the future locations, in the same way we have have perfect information about the locations that the algorithm already visited. This is an unrealistic assumption in practice, but it will help us to set-up our algorithm. We leave for the next section the details of how to marginalise over $\future_{n}$. 

Assume, for now, that $\future_{n}$ exists and that we have access to it. We it and denote by $\dataVector=(y_*,\dots,\dots,y_n)^T$ the vector of future locations evaluations of $f$ at $\future_n(\latentVector_*)$. Assuming that $\dataVector$ is known, it is possible to rewrite the expected loss in Eq.~(\ref{eq:expected_nonmyopic_loss}) as
\begin{equation}\label{eq:oracle_expected_nonmyopic_loss}
\Lambda_n \bigl(\latentVector_* \mid \I_0, \future_{n}(\latentVector_*) \bigr) = \E \bigl[\min (\dataVector,\eta) \bigr], 
\end{equation}
where the expectation is taken over the multivariate Gaussian distribution, with mean vector $\mu$ and covariance matrix $\Sigma$, that gives rise after marginalizing the posterior distribution of the GP at $\future_{n}(\latentVector_*)$. Note that under a fixed $\future_{n}(\latentVector_*)$, it also holds that 
$
\Lambda_n \bigl(\latentVector_* \mid \I_0, \future_{n}(\latentVector_*) \bigr) 
= 
\Gamma_n \bigl(\latentVector_* \mid \I_0,\future_{n}(\latentVector_*) \bigr)
$. 
See supplementary materials for details.

The intuition behind Eq.~(\ref{eq:oracle_expected_nonmyopic_loss}) is as follows: the expected loss at $\latentVector_*$ is the best possible function value that we expect to find in the next $n$ steps, conditional on the first evaluation being made at $\latentVector_*$. Rather than merely quantifying the benefit provided by the next evaluation, this loss function accounts for the expected gain in the whole future sequence of evaluations of $f$. As we analyse in the experimental section of this work, the effect of this is an adaptive loss that tends to be more explorative when more remaining evaluations are available and more exploitative as soon as we approach the final evaluations of $f$.

\begin{algorithm*}[t!]
   \caption{Decision process of the \us algorithm.}
   \label{alg:glasses}
\begin{algorithmic}
   \STATE {\bfseries Input:} dataset $\dataSet_{0} = \{(\textbf{x}_0, y_0)\}$, number of remaining evaluations ($n$), look-ahead predictor $\future$.
   \FOR{$j=0$ {\bfseries to} $n$ }
   \STATE 1. Fit a \gp with kernel $k$ to $\dataSet_{j}$.
   \STATE 2. Build a predictor of the future $n-l$ evaluations: $\hat{\future}_{n-j}$.
   \STATE 3. Select next location $\latentVector_j$ by taking  $\latentVector_j = \arg \min_{\latentVector \in {\mathcal X}}\Lambda_{n-j}(\latentVector_*|\I_0, \future_{n-j}(\latentVector_*))$.
   \STATE 4. Evaluate $f$ at $\latentVector_j$ and obtain $\dataScalar_j$.
   \STATE 5. Augment the dataset $\dataSet_{j+1} = \{\dataSet_{j} \cup (\textbf{x}_j, \dataScalar_j)\}$.
   \ENDFOR
   \STATE Fit a \gp with kernel $\kernelScalar$ to $\dataSet_{n}$
   \STATE \textbf{Returns}: Propose final location at $\arg \min_{\latentVector \in \inputSpace} \left\{\mu(\latentVector; \dataSet_{n})\right\}$.  
\end{algorithmic}
\end{algorithm*}

\begin{figure*}[t!]
\begin{tabular}{ccc}
      \addheight{\includegraphics[width=52mm]{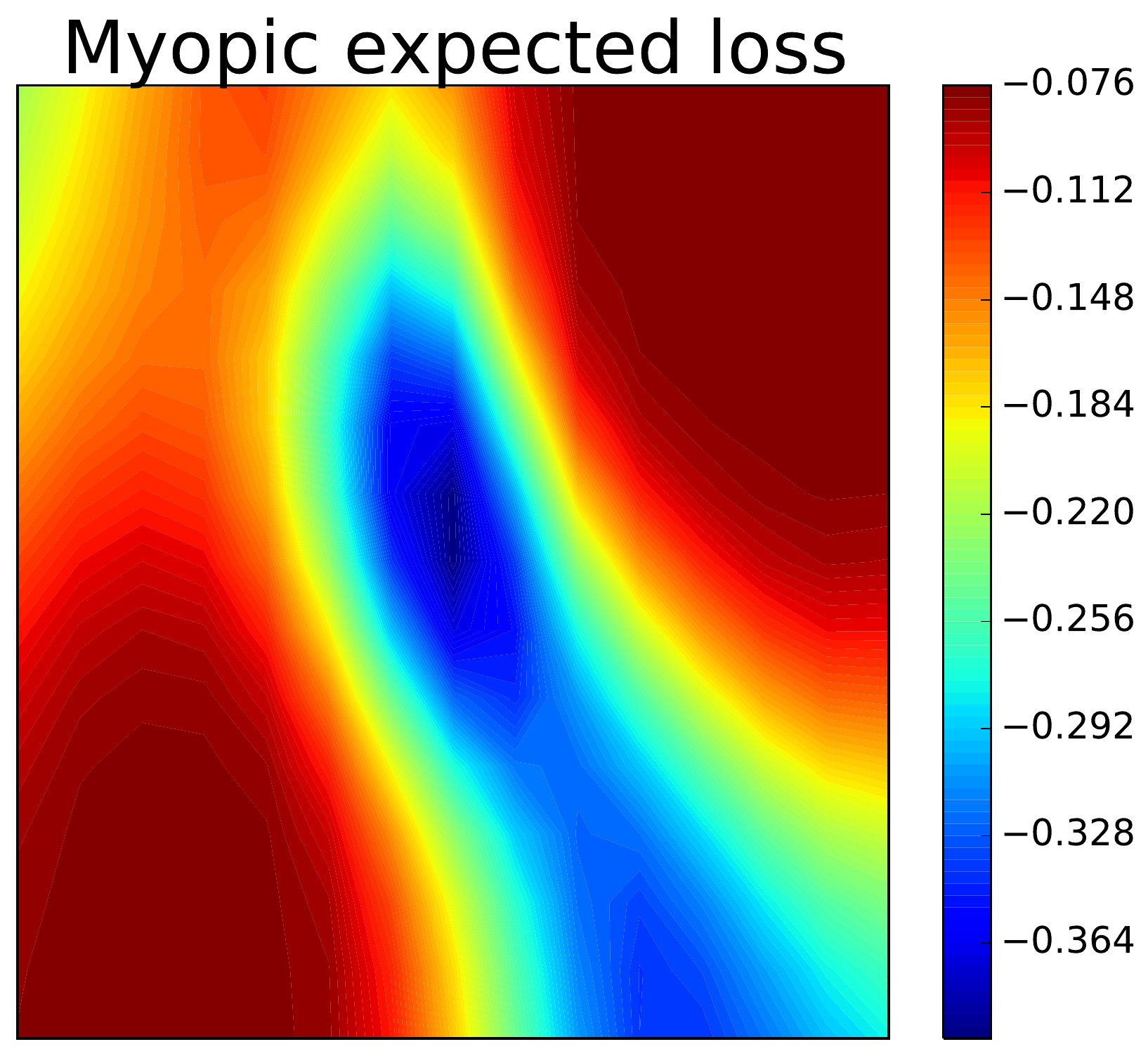}} &
      \addheight{\includegraphics[width=52mm]{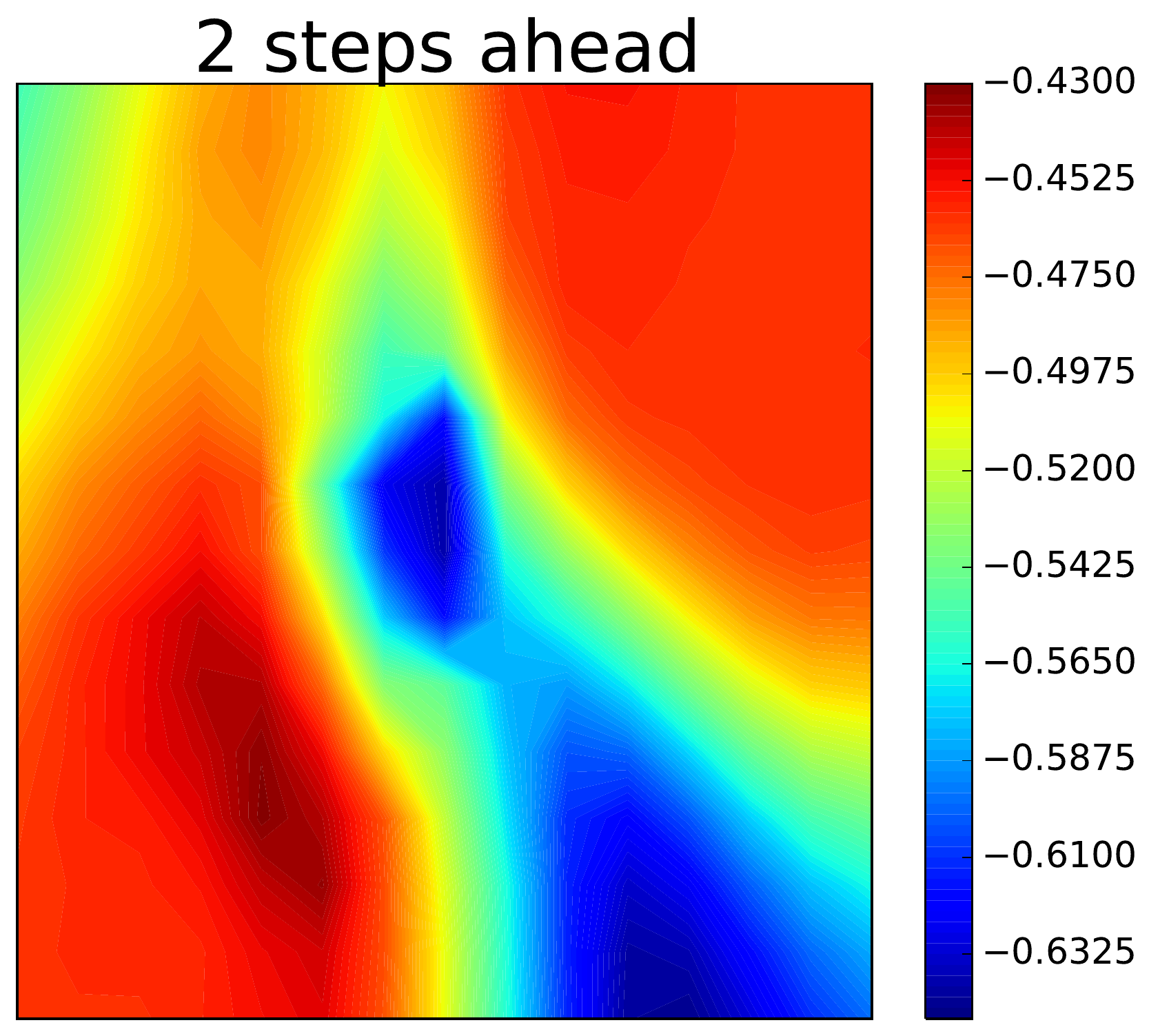}}  &
      \addheight{\includegraphics[width=52mm]{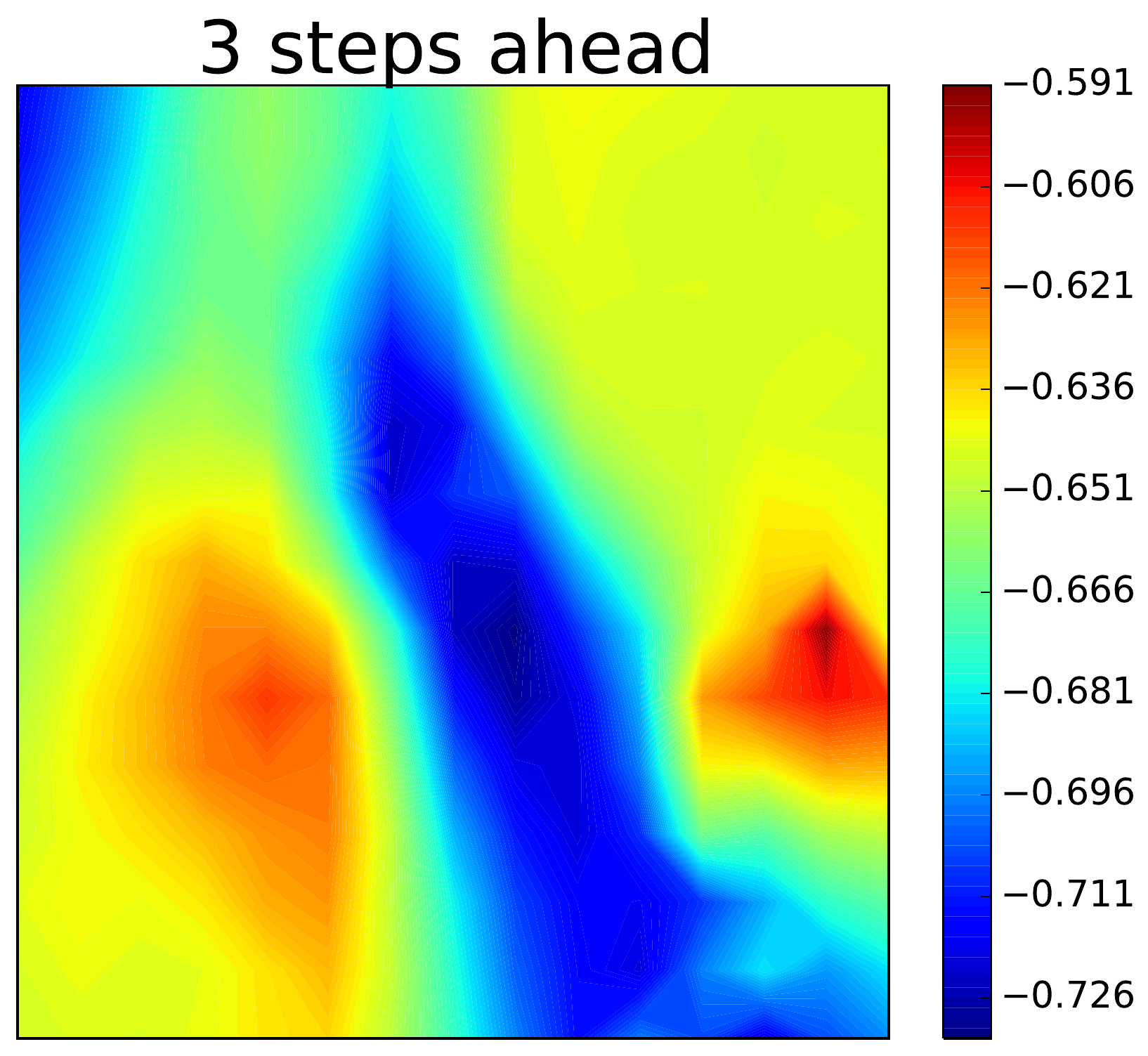}}\\
      \addheight{\includegraphics[width=52mm]{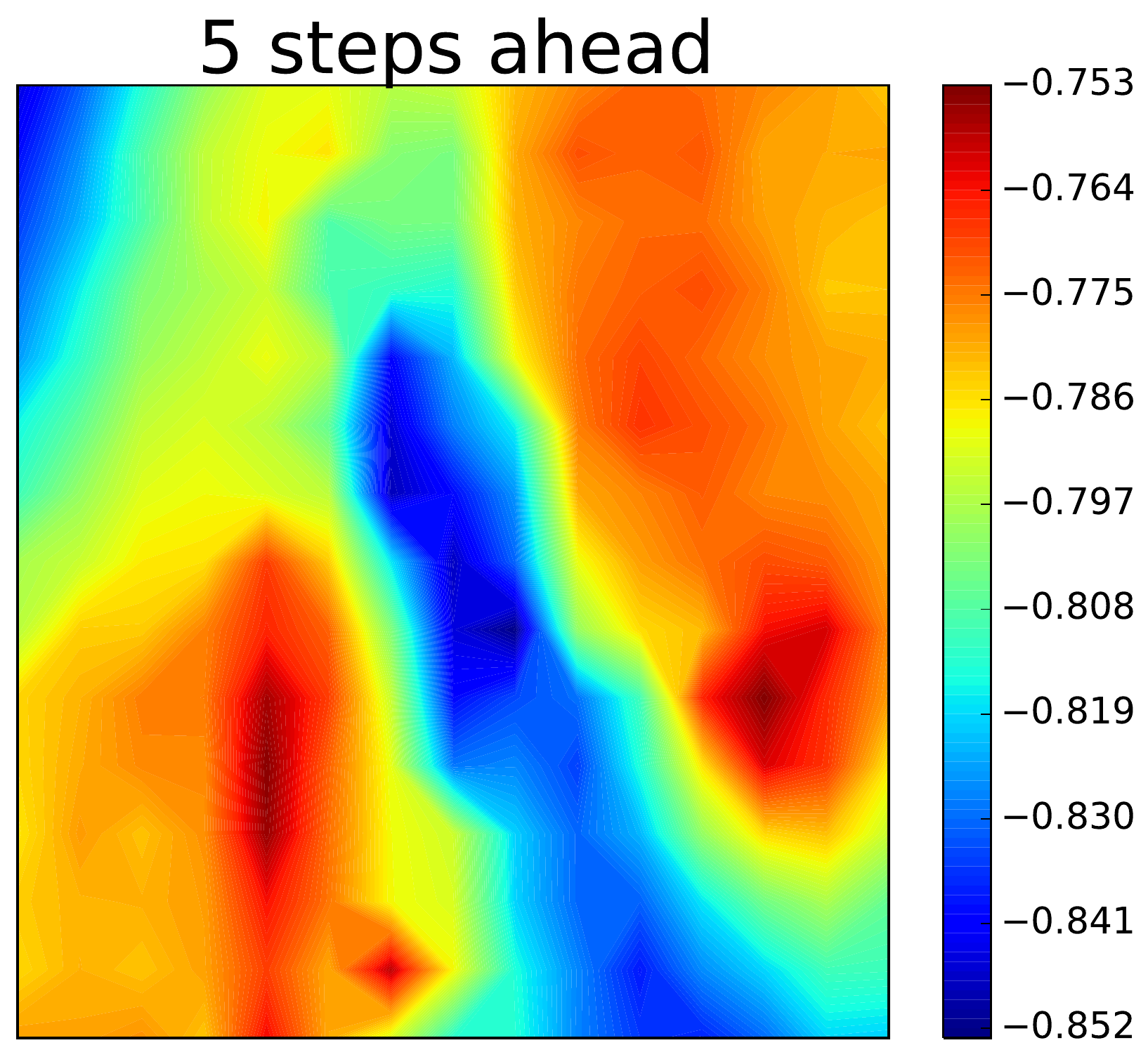}} &
      \addheight{\includegraphics[width=52mm]{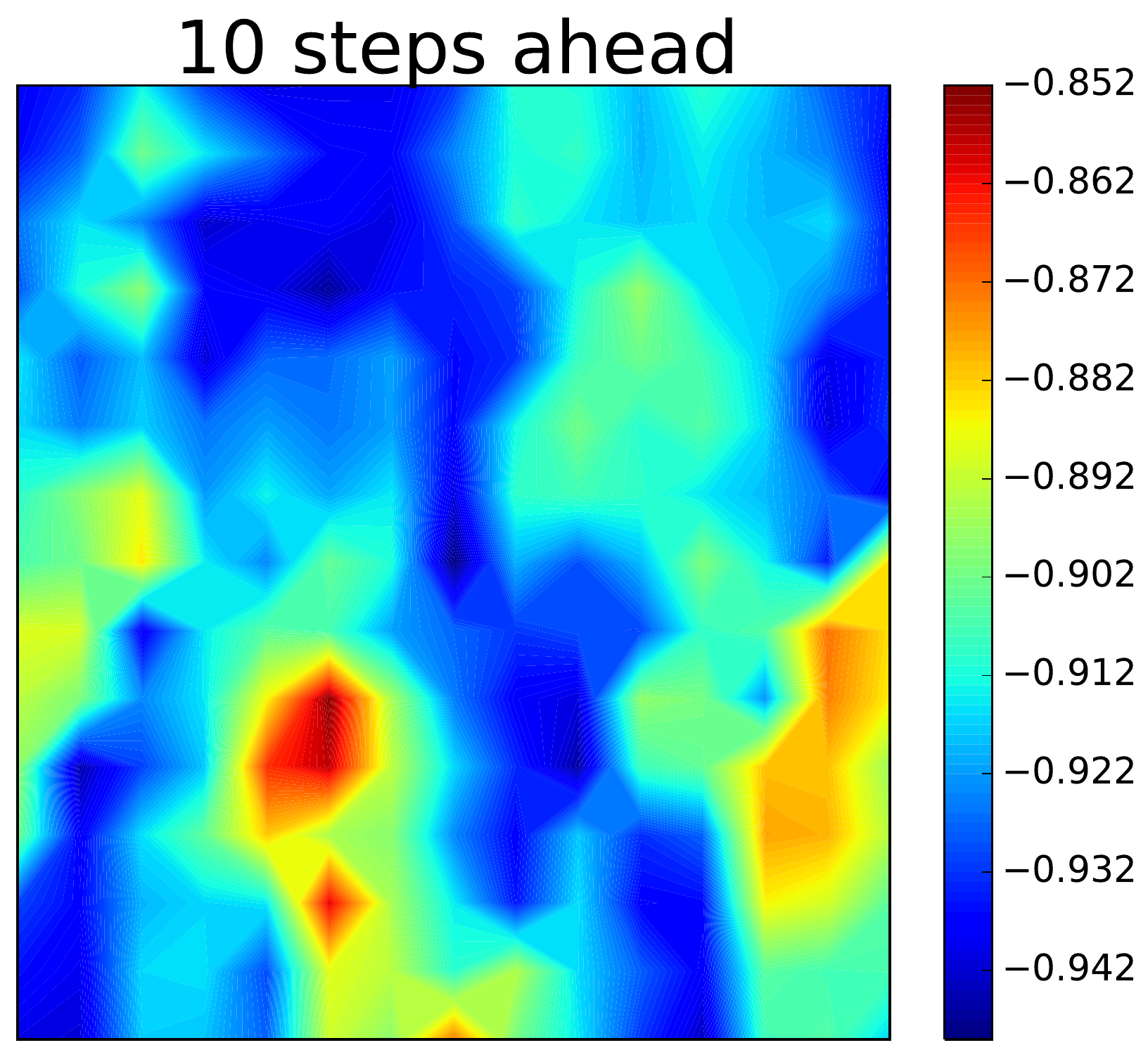}}  &
      \addheight{\includegraphics[width=52mm]{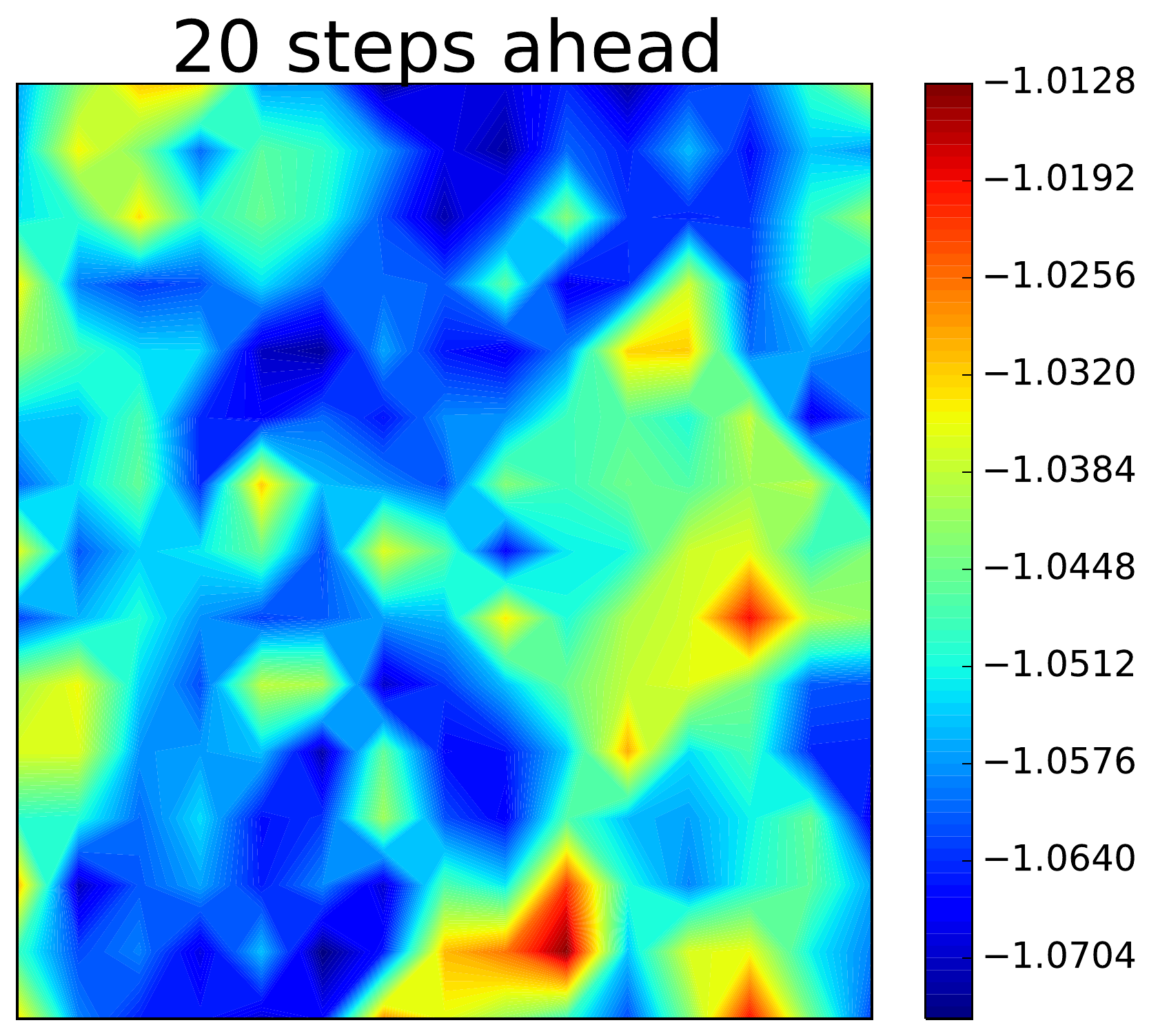}}\\
\end{tabular}\caption{Expected loss for different number of steps ahead in an example with 10 data points and the Six-hump Camel function. Increasing the number of steps-ahead flattens down the loss since it is likely for the algorithm to hit a good location irrespective of the initial point (all candidate points look better because of the future chances of the algorithm to be in a good location).}\label{table:n_ahead}
\end{figure*}

To compute Eq.~(\ref{eq:oracle_expected_nonmyopic_loss}) we use Expectation Propagation, \ep, \citep{Minka:2001}. This turns out to be a natural operation by observing that
\begin{eqnarray}\label{eq:expected_loss_oracle}
\E [\min (\dataVector,\eta)] & = & \eta\int_{\IR^n} \prod_{i=1}^nh_i(\dataVector) \N(\dataVector; \mu, \Sigma) \ud \dataVector \\  \nonumber
 &+ & \sum_{j=1}^n  \int_{\IR^n} \dataScalar_j \prod_{i=1}^n t_{j,i}(\dataVector) \N(\dataVector; \mu, \Sigma) \ud \dataVector
\end{eqnarray}
where  $h_i(\dataVector) = \mathbb{I}\{\dataScalar_i>\eta\}$ and
$$t_{j,i}(\dataVector)= \left\{ \begin{array}{lcl}
\mathbb{I}\{\dataScalar_j \leq\eta\} & \mbox{ if } $i=j$ \\
  \\
 \mathbb{I}\{ 0 \leq \dataScalar_i-\dataScalar_j \} &   \mbox{otherwise.} 
\end{array}
\right.$$
See supplementary materials for details. The first term in Eq.~(\ref{eq:expected_loss_oracle}) is a Gaussian probability on an unbounded polyhedron in which the limits are aligned with the axis. The second term is the sum of the Gaussian expectations on different non-axis-aligned different polyhedra defined by the indicator functions. Both terms can be computed with \ep using the approach proposed in \citep{Cunningham*Hennig*Lacoste-Julien_2011}. In a nutshell, to compute the integrals it is possible to replace the indicator functions with univariate Gaussians that play the role of \emph{soft-indicators} in the \ep iterations. This method is computationally efficient and scales well for high dimensions. Note that when $n=1$, Eq.~(\ref{eq:oracle_expected_nonmyopic_loss}) reduces to Eq.~(\ref{eq:expected_myopic_loss}).

Under the hypothesis of this section, the next evaluation is located where $\Lambda_n(\latentVector_*|\I_0, \future_{n}(\latentVector_*) )$ gives the minimum value. We still need, however, to propose a way to approximate the oracle $\hat{\future}_n(\latentVector_*)$. We do in next section.

\subsection{Local Penalisation to Predicting the Steps Ahead}



This section proposes an empirical surrogate $\hat{\future}_n(\latentVector_*)$ for $\future_n(\latentVector_*)$. A sensible option would be to use the \emph{maximum a posteriori probability}, \map, of the above-mentioned \dpp. However, as it is the generation of \dpp samples, to compute the \map of a \dpp is an expensive operation \citep{NIPS2012_4577}. Alternatively, here we use some ideas that have been recently developed in the batch Bayesian optimisation literature. In a nutshell, batch \bo methods aim to define sets of points in $\mathcal{X}$ where $f$ should be evaluated in parallel, rather than sequentially. In essence, a key aspect to building a `good' batch is the same as to computing a good approximation for $\Lambda_n(\latentVector_*|\I_0)$: to find a set of good locations at which to evaluate the objective. 

In this work we adapt to our context the batch \bo idea proposed by \cite{gonzalez2015batch}. Inspired by the repulsion properties of \dpp , \cite{gonzalez2015batch} propose to build each batch by iteratively optimising and penalising the acquisition function in a neighbourhood of the   already collected points by means of some local penalisers $\varphi(\latentVector;\latentVector_j)$. Note that any other batch method could be used here, but we consider this approach since it is computationally tractable and scales well with the size of the batches (steps ahead in our context) and the dimensionality of the problem.

More formally, assume that the objective function $f$ is $L$-Lipschitz continuous, that is, it satisfies that $|\latentForce(\latentVector_1) - \latentForce(\latentVector_2) | \leq L \|\latentVector_1 -\latentVector_2 \|$, $\forall\; \latentVector_1,\latentVector_2\in\mathcal{X}$. Take $M = \min_{\latentVector \in {\inputSpace}} \latentForce(\latentVector)$ and valid Lipschitz constant $L$. It is possible to show that  the ball
\begin{equation}\label{eq:radious}
B_{r_j}(\latentVector_j)= \{\latentVector\in \inputSpace : \|\latentVector_j-\latentVector \| \leq r_j \}
\end{equation}
where $r_j = \frac{\latentForce(\latentVector_j)-M}{L}$, doesn't contain the minimum of $f$. Probabilistic versions of these balls are used to define the above mentioned penalisers by noting that $\latentForce(\latentVector) \sim \GP(\mu(\latentVector),k(\latentVector,\latentVector'))$. In particular, $\varphi(\latentVector;\latentVector_j)$ is chosen as the probability that $\latentVector$, any point in $\inputSpace$ that is a potential candidate to be a minimum, does not belong to $B_{r_j}(\latentVector_j)$: $\varphi(\latentVector;\latentVector_j)  = 1- p (\latentVector  \in B_{r_j}(\latentVector_j)).$
As detailed in \cite{gonzalez2015batch}, these functions have a closed form and create an exclusion zone around the point $\latentVector_j$. The predicted k-th location when looking at $n$ step ahead and using $\latentVector_*$ as putative point is defined as
\begin{equation}\label{eq:penalized_acquisition}
 [\hat{\future}_n(\latentVector_*)]_k  =
 \arg \min_{x \in \inputSpace}\, \biggl\{g \bigl(\Lambda_1(\latentVector_* \mid \I_0) \bigr)\prod_{j=1}^{k-1}\varphi(\latentVector;\hat{\latentVector}_{j})\biggr\},
\end{equation}
for $k=2,\dots,n$ and where $\varphi(\latentVector;\latentVector_{j})$ are local local penalizers centered at $\latentVector_{j}$ and $g:\Re \rightarrow \Re^+$ is the \emph{soft-plus} transformation $g(z)= \ln(1+e^z)$.

To illustrate how $\hat{\future}_n$ computes the steps ahead we include Figure~(\ref{fig:steps_ahead}). We show the myopic loss in a \bo experiment together with predicted locations by $\hat{\future}_n$ for two different putative points. In case 1, the putative point $\latentVector_*$ (grey star) is close to the location of the minimum of the myopic loss (blue region). In case 2, $\latentVector_*$ is located in an uninteresting region. In both cases the future locations explore the interesting region determined for the myopic loss while the locations of the points are conditioned to the first location. In the bottom row of the figure we show how the points are selected in Case 1. The first putative point creates an exclusion zone that shifts the minimum of the acquisition, where the next location is selected. Iteratively, new locations are found by balancing diversity (due to the effect of the exclusion areas) and quality (exploring locations where the loss is low), similarly to the way samples the probabilities over subsets can be characterised in a \dpp \citep{MAL-044}.

\subsection{Algorithm and computational costs}
All the steps of \us are detailed in Algorithm \ref{alg:glasses}. The main computational cost is the calculation of the steps ahead, which is done using a sequence of \lbfgs optimizers at $\mathcal{O}(Pq^2)$ complexity for $P$, the maximum number of \lbfgs updates. The use of \ep to compute the value of the expected loss at each location requires a  quadratic run  time  factor  update in the dimensionality of each Gaussian factor.

\section{Experiments}\label{sec:experiments}


\subsection{Interpreting the non-myopic loss}
The goal of this experiment is to visualise the effect on the expected loss of considering multiple steps ahead. To this end, we use six-hump camel function (see Table~\ref{table:n_ahead} for details). We fit a \gp with a square exponential kernel and we plot the myopic expected loss together with 5 variants that consider 2, 3, 5, 10 and 20 steps ahead. Increasing the steps ahead decreases the optimum value of the loss: the algorithm can visit more locations and the expected minimum is lower. Also, increasing the steps ahead flattens down the loss: it is likely to hit a good location irrespective of the initial point so all candidate looks better because of the future chances of the algorithm to be in a good location. In practice this behaviour translates into an acquisition function that becomes more explorative as we look further ahead.

\subsection{Testing the effect of considering multiple steps ahead}

\begin{table}[t!]
\begin{center}
\begin{tabular}{lcc}
\toprule
Name &Function domain & $\inputDim$ \\
\midrule
SinCos & $[0,10]$ & 1 \\
Cosines & $[0,1]\times[0,1]$ & 2\\
Branin &$[-5,10]\times[-5,10]$ & 2\\
Sixhumpcamel  &$[-2,2]\times[-1,1]$ & 2\\
McCormick  & $[-1.5,4]\times[-3,4]$& 2\\
Dropwave   &$[-1,1]\times[-1,1]$ & 2\\
Beale   &$[-1,1]\times[-1,1]$ & 2\\
Powers  & $[-1,1]\times[-1,1]$& 2\\ 
Alpine2-$q$  & $[-10,10]^{\inputDim}$& 2, 5, 10\\
Ackley-$q$  &$[-5,5]^{\inputDim}$ & 2, 5\\
\bottomrule
\end{tabular}\caption{Details of the functions used in the experiments. The explicit form of these functions can be found at \url{http://www.sfu.ca/~ssurjano/optimization.html}, \citep{Molga1995} and the supplementary materials of this work. }\label{table:test_functions}
\end{center}
\end{table}

\begin{table*}[t!]
\begin{center}
\begin{tabular}{lcccccccc}
\toprule
{} &     MPI &     GP-LCB &      EL &    EL-2 &    EL-3 &    EL-5 &  EL-10 &    GLASSES \\
\midrule
SinCos  &  0.7147 &  0.6058 &  0.7645 &  \emph{0.8656} &  0.6027 &  0.4881 &  \emph{0.8274} &  \emph{\textbf{0.9000}} \\ 
Cosines           &  0.8637 &  0.8704 &  0.8161 &  \emph{0.8423} &  \emph{0.8118} &  0.7946 &  0.7477 &  \emph{\textbf{0.8722}} \\
Branin              &  0.9854 &  0.9616 &  \textbf{0.9900} &  0.9856 &  0.9673 &  0.9824 &  0.9887 &  0.9811 \\
Sixhumpcamel        &  0.8983 &  \textbf{0.9346} &  0.9299 &  0.9115 &  0.9067 &  0.8970 &  0.9123 &  0.8880 \\
Mccormick           & \textbf{0.9514} &  0.9326 &  0.9055 &  \emph{0.9139} &  \emph{0.9189} &  \emph{0.9283} &  \emph{0.9389} &  \emph{0.9424} \\
Dropwave            &  0.7308 &  0.7413 &  0.7667 &  0.7237 &  0.7555 &  0.7293 &  0.6860 &  \emph{\textbf{0.7740}} \\
Powers              &  0.2177 &  0.2167 &  0.2216 &  \emph{0.2428} &  \emph{0.2372} &  \emph{0.2390} &  \emph{0.2339} &  \emph{\textbf{0.3670}} \\
Ackley-2 &  0.8230 &  \textbf{0.8975} &  0.7333 &  0.6382 &  0.5864 &  0.6864 &  0.6293 &  0.7001 \\
Ackley-5  & 0.1832&   0.2082&   0.5473&   \emph{0.6694}&  0.3582&   0.3744&   \emph{\textbf{0.6700}} &  0.4348\\ 
Ackley-10 &  0.9893 &  0.9864 &  0.8178 &   \emph{0.9900} &   \emph{0.9912} &   \emph{\textbf{0.9916}} &   \emph{0.8340} &   \emph{0.8567} \\
Alpine2-2 &  \textbf{0.8628} &  0.8482 &  0.7902 &  0.7467 &  0.5988 &  0.6699 &  0.6393 &  0.7807 \\
Alpine2-5  &  0.5221 &  0.6151 &  \textbf{0.7797} &  0.6740 &  0.6431 &  0.6592 &  0.6747 &  0.7123 \\
\bottomrule
\end{tabular}\caption{Results for the average `gap' measure (5 replicates) across different functions.  \el-k is the expect loss function computed with $k$ steps ahead at each iteration. \us is the \us algorithm, \mpi is the maximum probability of improvement and \lcb is the lower confidence bound criterion. The best result for each function is bolded. In italic, the cases in which a non-myopic loss outperforms the myopic  loss are highlighted.}\label{table:comparision}
\end{center}
\end{table*}


To study the validity of our approximation we choose a variety of functions with a range of dimensions and domain sizes. See Table~\ref{table:test_functions} for details. We use the full \us algorithm (in which at each iteration the number of remaining iterations is used as the number of steps-ahead) and we show the results when 2, 3, 5, and 10 steps look-ahead are used to compute the loss function. Each problem is solved 5 times with different random initialisations of 5 data points. The number of allows evaluations is 10 times the dimensionality of the problem. This allows us to compare the average performance of each method on each problem. As baseline we use the myopic expected loss, \el. For comparative purposes we used other two loss functions are commonly used in the literature. In particular, we use the Maximum Probability of Improvement, \mpi, and the \gp lower confidence bound, \lcb. In this later case we set the parameter that balances exploration and exploitation to 1. See \citep{Snoek*Larochelle*Adams_2012} for details on these loss functions. All acquisition functions are optimised using the dividing rectangles algorithm \direct \citep{Jones1993}.
 As surrogate model for the functions we used a \gp with a squared exponential kernel plus a bias kernel \citep{Rasmussen:2005:GPM:1162254}. The hyper-parameters of the model were optimised by the standard method of maximising the marginal likelihood, using \lbfgs \citep{Nocedal1980} for 1,000 iterations and selected the best of 5 random restarts.
To compare the methods we used the `gap' measure of performance \citep{Huang:2006}, which is defined as
$$G \triangleq \frac{y(\latentVector_{first})-y(\latentVector_{best})}{y(\latentVector_{first})-y(\latentVector_{opt})},$$
where $y(\cdot)$ represents the evaluation of the objective function, $y(\latentVector_{opt})$ is the global minimum, and $\latentVector_{first}$ and $\latentVector_{best}$ are the first and best evaluated point, respectively. To avoid gap measures larger that one due to the noise in the data, the measures for each experiment are normalized across all methods. The initial point $\latentVector_{first}$ was chosen to be the best of the original points used to initialise the models. 

Table \ref{table:comparision} shows the comparative across different functions and methods. None of the methods used is universally the best but a non myopic loss is the best in 6 of the 11 cases. In 3 cases the full \us approach is the best of all methods. Specially interesting is the case of the McCormick and the Powers function. In these two cases, to increase the number of steps ahead used to compute the loss consistently improve the obtained results. Note as well that when the \us algorithm is not the best global method it typically performs closely to the best alternative which makes it a good `default' choice if the function to optimise is expensive to evaluate.



\section{Conclusions}\label{sec:conclusions}

In this paper we have explored the myopia in Bayesian optimisation methods. For the first time in the literature, we have proposed an non-myopic loss that allows taking into account dozens of future evaluations before making the decision of where to sample the objective function. The key idea is to jointly model all future evaluations of the algorithm with a probability distribution and to compute the expected loss by marginalising them out. Because this is an expensive step, we avoid it by proposing a fixed prediction of the future steps. Although this doesn't make use of the epistemic uncertainty on the steps ahead, it drastically reduces the computation burden of approximating the loss. We made use of the connection of the multiple steps ahead problem with some methods proposed in the batch Bayesian optimisation to solve this issue. The final computation of the loss for each point in the domain is carried out by adapting \ep to our context. As previously suggested in \cite{osborne_gaussian_2009}, our results confirm that using a non-myopic loss helps, in practice, to solve global optimisation problems. Interestingly, and as happens with any comparison of loss functions across many objective functions, there is not a universal best method. However, in cases in which \us is not superior, it performs very closely to the myopic loss, which makes it an interesting default choice in most scenarios. 

Some interesting challenges will be addressed in the future such as making the optimisation of the loss more efficient (for which \direct is employed in this work): although the smoothness of the loss is guaranteed if the steps ahead are consistently predicted for points close in the space, if the optimization of the steps ahead fails,  the optimization of the loss may be challenging.  Also, the use of non-stationary kernels, extensions to deal with to high dimensional problems and finding efficient was of sampling many steps ahead will also be analysed.

\bibliographystyle{plainnat}
\bibliography{bib_glasses}

\newpage
\clearpage
\setcounter{section}{0}
\setcounter{equation}{0}
\renewcommand{\thesection}{S\arabic{section}}
\renewcommand{\theequation}{S.\arabic{equation}}
\onecolumn
\begin{center}
{\Large  \textbf{Supplementary materials for:\\
`GLASSES: Relieving The Myopia Of Bayesian Optimisation'}}
\end{center}


\vspace{1cm}


\section{Oracle Multiple Steps look-ahead Expected Loss }
Denote by $\eta_n = \min \{\dataVector_0, y_*, y_2\dots,y_{n-1}\}$ the value of the best visited location when looking at $n$ evaluations in the future. Note that $\eta_n$ reduces to the current best lost $\eta$ in the one step-ahead case. It is straightforward to see that 
$$ \min (y_n,\eta_n) = \min (\dataVector,\eta ).$$
It holds hat
\begin{eqnarray}\nonumber
\Lambda_n(\latentVector_*|\I_0, \future_{n}(\latentVector_*)) & = & \int \min (\dataVector,\eta) \prod_{j=1}^{n}p(y_{j}|\I_{j-1}, \future_{n}(\latentVector_*)) \ud y_*\dots \ud y_n
\end{eqnarray}
where the integrals with respect to $\latentVector_2\dots \ud\latentVector_n$ are  $p(\latentVector_{j}|\I_{j-1}, \future_{n}(\latentVector_*))=1$, $j=2,\dots,n$ since we don't need to optimize for any location and $p(y_{j}|\latentVector_{j},\I_{j-1}, \future_{n}(\latentVector_*))=p(y_{j}|\I_{j-1}, \future_{n}(\latentVector_*))$. Notice that
\begin{eqnarray}\nonumber
\prod_{j=1}^{n}p(y_{j}|\I_{j-1}, \future_{n}(\latentVector_*))& =& p(y_n|\I_{n-1}, \future_{n}(\latentVector_*)) \prod_{j=1}^{n-1}p(y_{j}|\I_{j-1} \future_{n}(\latentVector_*))\\\nonumber
& = & p(y_n,y_{n-1}|\I_{n-2}, \future_{n}(\latentVector_*))  \prod_{j=1}^{n-2}p(y_{j}|\I_{j-1} \future_{n}(\latentVector_*))\\\nonumber
& & \dots \\\nonumber
& = & p(y_n,y_{n-1},\dots,y_2|\I_{1}, \future_{n}(\latentVector_*))\prod_{j=1}^{2}p(y_{j}|\I_{j-1} \future_{n}(\latentVector_*))\\\nonumber
& = & p(\dataVector|\I_{0}, \future_{n}(\latentVector_*)) \nonumber
\end{eqnarray}
and therefore 
$$ \Lambda_n(\latentVector_*|\I_0, \future_{n}(\latentVector_*)) =\E [\min (\dataVector,\eta)] =\int \min (\dataVector,\eta)p(\dataVector|\I_{0}, \future_{n}(\latentVector_*))d\dataVector  $$

\vspace{1cm}
\section{Formulation of the Oracle Multiple Steps loook-ahead Expected Loss to be computed using Expectation Propagation}
Assume that $\dataVector \sim \N(\dataVector; \mu, \Sigma)$. Then we have that
\begin{eqnarray}\nonumber
\E[\min (\dataVector,\eta)] & = & \int_{\IR^n} \min (\dataVector,\eta)  \N(\dataVector; \mu, \Sigma) d\dataVector\\ \nonumber
& = & \int_{\IR^n - (\eta,\infty)^n } \min (\dataVector)  \N(\dataVector; \mu, \Sigma) d\dataVector + \int_{(\eta,\infty)^n} \eta  \N(\dataVector; \mu, \Sigma) d\dataVector.  \nonumber
\end{eqnarray}
The first term can be written as follows:
\begin{equation}
 \int_{\IR^n - (\eta,\infty)^n } \min (\dataVector)  \N(\dataVector; \mu, \Sigma) d\dataVector  =    \sum_{j=1}^n \int_{P_j} \dataScalar_j \N(\dataVector; \mu, \Sigma) \ud \dataVector \nonumber
\end{equation}\nonumber
where $P_j := \{ \dataVector \in\IR^n - (\eta,\infty)^n  : \dataScalar_j \leq \dataScalar_i,\,\, \forall i \neq j \}$. We can do this because the regions $P_j$ are disjoint and it holds that $\cup_{j=1}^{n}P_j = \IR^n - (\eta,\infty)^n $.  Also, note that the $\min(\dataVector)$ can be replaced within the integrals since within each $P_j$ it holds that $\min(\dataVector) = \dataScalar_j$. Rewriting the integral in terms of indicator functions we have that
\begin{eqnarray}\label{eq:term1}
 \sum_{j=1}^n \int_{P_j} \dataScalar_j \N(\dataVector; \mu, \Sigma) \ud \dataVector   =  \sum_{j=1}^n  \int_{\IR^n} \dataScalar_j \prod_{i=1}^n t_{j,i}(\dataVector) \N(\dataVector; \mu, \Sigma) \ud \dataVector 
\end{eqnarray}

where $t_{j,i}(y) =\mathbb{I}\{\dataScalar_i \leq\eta\}$ if $j=i$ and $t_{j,i}(y) =\mathbb{I}\{\dataScalar_j \leq \dataScalar_i \}$ otherwise.

The second term can be written as
\begin{equation}\label{eq:term2}
 \int_{(\eta,\infty)^n } \eta  \N(\dataVector; \mu, \Sigma) d\dataVector = \eta\int_{\IR^n} \prod_{i=1}^nh_i(\dataVector) \N(\dataVector; \mu, \Sigma) d\dataVector
\end{equation}
where $h_i(\dataVector) = \mathbb{I}\{\dataScalar_i>\eta\}$.  Merge (\ref{eq:term1}) and (\ref{eq:term2}) to obtain Eq.  (\ref{eq:expected_loss_oracle}).
 
\subsection{Synthetic functions}

In this section we include the formulation of the objective functions used in the experiments that are not available in the references provided.

\begin{table}[h!]
\centering
\begin{tabular}{cc}
\toprule
Name & Function     \\
\midrule
SinCos & $f(\latentScalar) = \latentScalar \sin(\latentScalar) + \latentScalar \cos(2\latentScalar)$  \\
Alpine2-$\inputDim$ & $f(\latentVector) =\prod_{i=1}^{\inputDim} \sqrt{\latentScalar_i}\sin(\latentScalar_i)$  \\
Cosines &  $f(\latentVector) = 1- \sum_{i=1}^2 (g(\latentScalar_i) - r(\latentScalar_i) )  \mbox{ with } g(\latentScalar_i) = (1.6\latentScalar_i - 0.5)^2 \mbox{ and } r(\latentScalar_i) = 0.3 \cos (3\pi (1.6 \latentScalar_i -0.5))$. \\
\bottomrule
\end{tabular}\caption{Functions used in the experimental section.}\label{table:functions_test}
\end{table}


\end{document}

%% file: notationDef.tex
\global\long\def\inputSpace{\mathcal{X}}

\global\long\def\inputDim{q}

\global\long\def\dataStd{\sigma}

\global\long\def\parameterScalar{\theta}
\global\long\def\parameterVector{\boldsymbol{\parameterScalar}}

\global\long\def\dataScalar{y}
\global\long\def\dataVector{\mathbf{\dataScalar}}

\global\long\def\dataSet{\mathcal{D}}

\global\long\def\latentScalar{x}
\global\long\def\latentMatrix{\mathbf{\MakeUppercase{\latentScalar}}}
\global\long\def\latentVector{\mathbf{\latentScalar}}

\global\long\def\noiseScalar{\epsilon}

\global\long\def\kernelScalar{k}
\global\long\def\kernelVector{\mathbf{\kernelScalar}}
\global\long\def\kernelMatrix{\mathbf{\MakeUppercase{\kernelScalar}}}

\global\long\def\latentForce{f}

\global\long\def\parameterScalar{\theta}
\global\long\def\parameterVector{\boldsymbol{\parameterScalar}}

